%% file: main.tex
\documentclass{article}

 \usepackage[dblblindworkshop, final]{neurips_2025}
\workshoptitle{What Makes a Good Video:
Next Practices in Video Generation and Evaluation}

\usepackage{graphicx} 
\usepackage{amsmath}
\usepackage{amsthm}
\usepackage{algorithm}
\usepackage{algpseudocode}
\usepackage{amsfonts}
\usepackage{graphicx}
\usepackage{xcolor}

\usepackage[utf8]{inputenc} 
\usepackage[T1]{fontenc}    
\usepackage{hyperref}       
\usepackage{url}            
\usepackage{booktabs}       
\usepackage{amsfonts}       
\usepackage{nicefrac}       
\usepackage{microtype}      
\usepackage{xcolor}         

\title{VideoGen-of-Thought: Step-by-step generating multi-shot video with minimal manual intervention}

\author{%
  Mingzhe Zheng\textsuperscript{1,6}
  Yongqi Xu\textsuperscript{2,6}
  Haojian Huang\textsuperscript{3} 
  Xuran Ma\textsuperscript{1,6}
  Yexin Liu\textsuperscript{1,6}
  Wenjie Shu\textsuperscript{1,6} \And 
  Yatian Pang\textsuperscript{4,6} 
  Feilong Tang\textsuperscript{1,6} 
  Qifeng Chen\textsuperscript{1,†}
  Harry Yang\textsuperscript{1,6,†}
  Ser-Nam Lim\textsuperscript{5,6,†}
  \\
  \\
  \textsuperscript{1} Hong Kong University of Science and Technology \quad 
  \textsuperscript{2} Peking University \quad \\
  \textsuperscript{3} University of Hong Kong \quad 
  \textsuperscript{4} NUS \quad
  \textsuperscript{5} University of Central Florida \quad
  \textsuperscript{6} Everlyn AI \quad
}

\begin{document}

\maketitle

\begin{figure*}[!h]
	\centering
	\begin{tabular}{cc}
	\hspace{-1.5cm}
	\includegraphics[width = 1.0\linewidth]{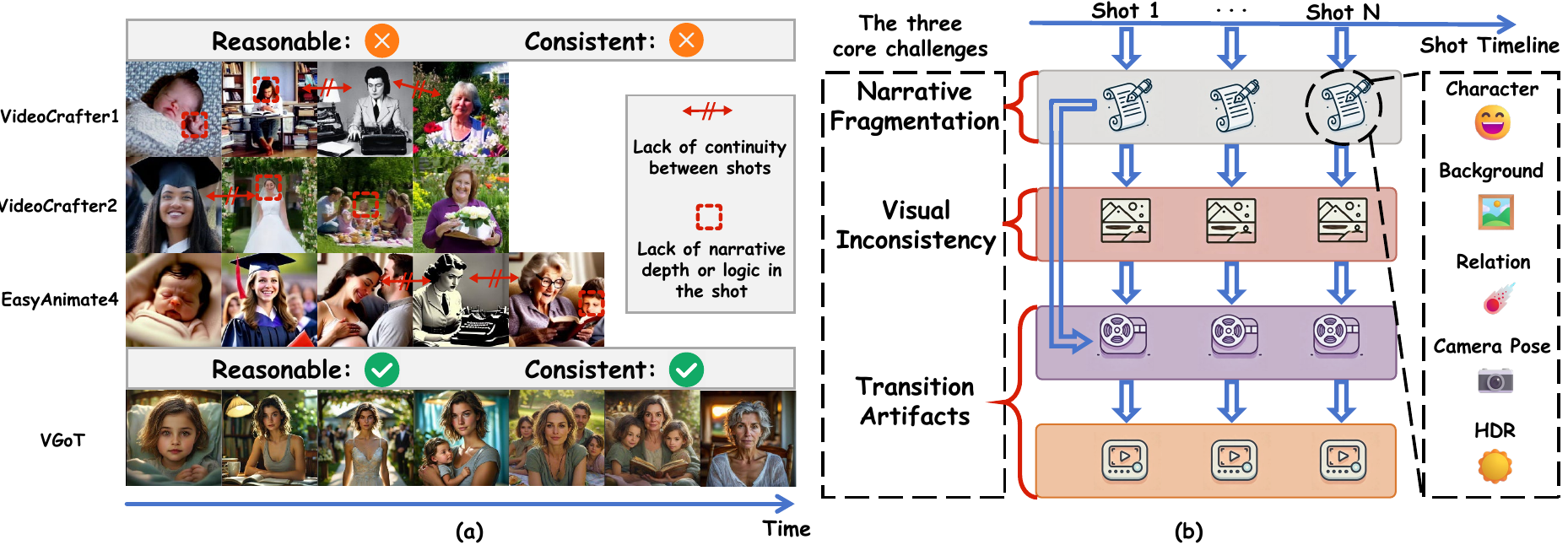}
	\end{tabular}
	\caption{\textbf{Illustration of \textit{VideoGen-of-Thought (VGoT)}.} \textbf{(a) Comparison of existing methods with \textit{VGoT} in multi-shot video generation.} 
   Existing methods struggle with maintaining reasonability and consistency across multiple shots, while \textit{VGoT} effectively addresses these challenges through a multi-shot generation approach. 
   \textbf{(b) Challenges solved by \textit{VGoT}}: addressing \textit{narrative fragmentation} with dynamic storylines modeling across five domains \textit{(characters/backgrounds/relations/camera/HDR)}, tackling \textit{visual inconsistency} via identity-aware cross-shot propagation to create keyframes using IPP tokens derived from narrative elements, and solving \textit{transition artifacts} during multi-shot video synthesizes through adjacent latent transition mechanisms.}
	\label{fig:1}
\end{figure*}

\begingroup
\renewcommand{\thefootnote}{}
\footnotetext{†~Corresponding author.}
\footnotetext{Project webpage: \href{https://cheliosoops.github.io/VGoT/}{https://cheliosoops.github.io/VGoT/}}
\endgroup

\input{sec/0_abstract}

\input{sec/1_intro}
\input{sec/2_related_works}
\input{sec/3_Preliminaries}

\input{sec/4_Methods}
\input{sec/5_experiments}
\input{sec/6_conclusion}
{
    \small
    \bibliographystyle{ieeenat_fullname}
    \bibliography{reference}
}


\appendix

\input{sec/X_suppl}


\newpage
\input{sec/checklist}

\end{document}

%% file: sec/0_abstract.tex
\begin{abstract}
Current video generation models excel at short clips but fail to produce cohesive multi-shot narratives due to disjointed visual dynamics and fractured storylines. 
Existing solutions either rely on extensive manual scripting/editing or prioritize single-shot fidelity over cross-scene continuity, limiting their practicality for movie-like content. 
We introduce \textbf{VideoGen-of-Thought (VGoT)}, a step-by-step framework that automates multi-shot video synthesis \textbf{from a single sentence} by systematically addressing three core challenges: 
\textbf{(1) Narrative fragmentation}: Existing methods lack structured storytelling.
We propose dynamic storyline modeling, which turns the user prompt into concise shot drafts and then expands them into detailed specifications across five domains (character dynamics, background continuity, relationship evolution, camera movements, and HDR lighting) with self-validation to ensure logical progress.
\textbf{(2) Visual inconsistency}: previous approaches struggle to maintain consistent appearance across shots.
Our identity-aware cross-shot propagation builds identity-preserving portrait (IPP) tokens that keep character identity while allowing controlled trait changes (expressions, aging) required by the story.
\textbf{(3) Transition artifacts}: Abrupt shot changes disrupt immersion.
Our adjacent latent transition mechanisms implement boundary-aware reset strategies that process adjacent shots' features at transition points, enabling seamless visual flow while preserving narrative continuity.
Combined in a training-free pipeline, VGoT surpasses strong baselines by 20.4\% in within-shot face consistency and 17.4\% in style consistency, while requiring \textbf{10× fewer} manual adjustments.
VGoT bridges the gap between raw visual synthesis and director-level storytelling for automated multi-shot video generation.

\end{abstract}

%% file: sec/1_intro.tex
\section{Introduction}
\label{sec:intro}

Recent advancements in video generation techniques have yielded impressive results, particularly in creating short, visually appealing clips~\cite{blattmann2023stable, chen2023videocrafter1, chen2024videocrafter2, xu2024easyanimate, henschel2024streamingt2v}.
These advancements have been powered by increasingly sophisticated generative models, ranging from diffusion models~\cite{ho2020denoising, song2020scorebased, rombach2022stablediffusion, blattmann2023stable} to auto-regressive models~\cite{ge2022tats, weng2023artboldsymbolcdotv, liu2024mardini, wang2024omnitokenizer}, supported by large-scale datasets~\cite{huang2020movienet, schuhmann2021laion400m, schuhmann2022laion}.
These methods have enabled the generation of high-quality and realistic short videos.
However, generating multi-shot videos from a brief user input script remains a substantial challenge. 
Unlike single-shot video generation, which focuses on creating a coherent clip from a single prompt, multi-shot video generation requires the model to maintain both \textbf{reasonable storylines} and \textbf{visual consistency} across multiple shots. 
This task involves additional complexities, such as ensuring logical transitions between scenes and maintaining consistent appearance \textit{(e.g., character identity, overall style, etc.)} throughout the video. 
Current video generation methods~\cite{chen2023videocrafter1, chen2024videocrafter2, xu2024easyanimate, hong2022cogvideo, yang2024cogvideox} often fall short in these areas, resulting in fragmented narratives and inconsistent visual elements across shots. 
Furthermore, sometimes real movies require the same character to appear in different ways based on the storylines, which should be faithful to the same identity but not the same traits \textit{(e.g., expression, appearance, relationships, etc.).} The requirement for high-level identity preservation~\cite{ye2023ip-adapter, yuan2024consisid, zhou2025storydiffusion} across shots remains an open question, which is essential for cross-shot consistency.

Existing multi-shot video generation approaches suffer from several limitations. 
For instance, methods like MovieDreamer~\cite{zhao2024moviedreamer} require plenty of manual input, including mountains of script writing \textit{(e.g., character appearance, scene elements, detailed plots, etc.)} and image selection.
Other approaches, such as DreamFactory~\cite{xie2024dreamfactory}, focus on multi-agent pipelines but require specific documents and repeated manual adjustment for each story, restricting the capability of easy usage.
The need for heavy manual intervention not only increases the workload but also limits their practicality for automated movie-like content creation.
In contrast, our work aims to decompose the complex task of multi-shot video generation into smaller, manageable problems and solve them in a step-by-step manner with minimal manual intervention, as shown in Figure~\ref{fig:1}.
Our approach proposes to systematically and automatically address three core challenges: \textbf{(1) narrative fragmentation} through dynamic storyline modeling; \textbf{(2) visual inconsistency} via identity-aware cross-shot propagation; and \textbf{(3) transition artifacts} using adjacent latent transition mechanisms.


We propose \textbf{VideoGen-of-Thought (\textit{VGoT})}, an end-to-end framework that generates multi-shot video with \textit{reasonable storylines} and \textit{visual consistency} \textbf{from one sentence} with \textbf{minimal manual intervention}. 
Our framework addresses three fundamental challenges through a structured pipeline as shown in Fig~\ref{fig:2}:
First, \textit{VGoT} tackles \textit{narrative fragmentation} through converting a brief user prompt into short descriptions for across shots to obtain a reasonable story draft: we introduce a dynamic storyline modeling that transforms user prompts into shot sequences through a two-stage process with self-validation mechanisms that enforce narrative coherence by rejecting candidates violating cinematic principles. 

Additionally, to resolve \textit{visual inconsistency}, we introduce identity-aware cross-shot propagation that extracts multi-aspect portrait schemata from narrative elements, ranging from different avatars in the same story to different traits of the same identity following the development of the story. 
This system handles both inter-story avatar variations and intra-story identity evolution, generating identity-preserving portrait (IPP) tokens.
These IPP tokens guide keyframe generation through hierarchical feature injection in pretrained diffusion models~\cite{song2020ddim,rombach2022stablediffusion,kolors,ye2023ip-adapter}, ensuring style uniformity and identity fidelity across shots.

We encode each keyframe into latent space and utilize a video diffusion model to refine a noise map into video latent codes, representing $k$ frames of shot video, conditioned on the keyframe latent and the corresponding textual latent. To address \textit{transition artifacts}, we design a cross-shot transition mechanism with a FIFO-like~\cite{kim2024fifo} latent reset strategy that processes adjacent shots' features at the boundary, ensuring seamless transitions and maintaining visual continuity across shots while preserving the logical coherence established in the storyline preparation.


Additionally, current evaluation protocols for multi-shot video generation remain inadequate due to the absence of dedicated datasets and task-specific metrics. To address this limitation, we propose four novel quantitative metrics for systematic assessment:

\begin{itemize}
    \item \textbf{Within-Shot Face Consistency (WS-FC):} facial similarity between frames in a single shot.
    \item \textbf{Cross-Shot Face Consistency (CS-FC):} identity distance across shots.
    \item \textbf{Within-Shot Style Consistency (WS-SC):} style similarity between frames in a single shot.
    \item \textbf{Cross-Shot Style Consistency (CS-SC):} stylized bias across shots.
\end{itemize}

Experimental results demonstrate VGoT's superiority over state-of-the-art (SOTA) methods across all metrics. Quantitative comparisons reveal \textbf{20.4\%} and \textbf{17.4\%} improvements in WS-FC and WS-SC, respectively, compared to previous SOTA baselines. For cross-shot metrics, VGoT achieves \textbf{2.9$\times$} higher in CS-FC and \textbf{106.6\%} higher in CS-SC over baselines.
Human evaluations (Table~\ref{tab:2}) confirm these findings, with VGoT receiving \textbf{66.7\%} "Good" ratings for cross-shot consistency versus \underline{27.2\%} for competitors.


The principal contributions of this work can be summarized as follows:

\begin{itemize}
     
\item\textbf{Automated Multi-Shot Generation Framework:} We present VGoT, the first end-to-end system that generates story-coherent multi-shot videos from single sentence inputs, obviously reducing manual intervention.
    
\item\textbf{Three-Core Solution Architecture:} We propose a structured four-module pipeline addressing \textit{narrative fragmentation} through LLM-powered story decomposition, \textit{visual inconsistency} via identity-aware propagation, and \textit{transition artifacts} using adjacent latent transition mechanisms.
    
\item\textbf{Multi-Shot Evaluation Protocol:} We design a new multi-shot video assessment protocol featuring hierarchical consistency measurement with four quantitative metrics covering face/style consistency across both within-shot and cross-shot.

\end{itemize}

%% file: sec/2_related_works.tex
\section{Related Work}
\label{sec:related}


Video generation has made significant strides following the great success of diffusion models, leading to two important research categories: long video synthesis and multi-shot story generation. These areas focus on generating high-quality, consistent videos either as extended single shots or as coherent sequences across multiple scenes.

\noindent \textbf{Long Video Synthesis.}
Long video synthesis has advanced through diffusion-based methods and autoregressive approaches. Diffusion models based on Stable Diffusion~\cite{rombach2022stablediffusion} utilize iterative refinement to generate visually consistent frames and have been effective for short sequences~\cite{he2022lvdm,blattmann2023stable, chen2023videocrafter1, chen2024videocrafter2, xing2025dynamicrafter,wu2023tune,blattmann2023align,guo2023animatediff,yang2024cogvideox,zhang2024show,geyer2023tokenflow,huang2024vbench,peng2024controlnext,esser2023structure,ho2022video,singer2022make,zeng2024make,zhou2022magicvideo,qiu2023freenoise,pan2024synthesizing}. However, they struggle with maintaining coherence over extended video lengths. Autoregressive methods~\cite{ge2022tats, li2024arlon,yin2023nuwa} predict frames sequentially but often face error accumulation, making long-term consistency challenging and computationally expensive. Additionally, training-free methods like FIFO-Diffusion~\cite{kim2024fifo} generate long sequences without training but lack mechanisms to manage transitions across shots, limiting their effectiveness in narrative-driven content. Overall, while these approaches achieve visual fidelity, they fail to ensure logical coherence across extended sequences. In contrast, VideoGen-of-Thought (\textit{VGoT}) leverages a modular approach that includes cross-shot smoothing mechanisms to ensure both visual consistency and narrative coherence, offering a more holistic solution for generating long-form videos.

\noindent\textbf{Multi-Shot Video Generation.} Multiple Shot Story Generation focuses on maintaining narrative coherency across distinct scenes, and existing approaches face critical limitations in automation scalability. Animate-a-Story~\cite{he2023animate} uses retrieval-augmented generation to ensure visual consistency but struggles with maintaining logical narrative transitions. MovieDreamer~\cite{zhao2024moviedreamer} requires extensive manual scripting \textit{(character profiles, scene details, etc.)} and image curation. DreamFactory~\cite{xie2024dreamfactory} demands repetitive adjustment and special documents for each story in its multi-agent system across stories. Flexifilm~\cite{ouyang2024flexifilm} and StoryDiffusion~\cite{zhou2025storydiffusion} introduce conditional adaptability but still necessitate manual consistency fixes between shots. These methods share a common bottleneck: heavy reliance on human intervention for narrative and visual coherence. 
\textit{VGoT} breaks this paradigm through systematic automation: 1) \textit{Narrative Fragmentation} → LLM-driven story decomposition; 2) \textit{Visual Inconsistency} → Story-derived identity propagation; 3) \textit{Transition Artifacts} → Boundary-aware latent processing. This structured approach enables director-level storytelling from single prompts with 10× fewer manual interventions than prior works~\cite{zhao2024moviedreamer,xie2024dreamfactory}.



%% file: sec/3_Preliminaries.tex
\section{Preliminaries and Problem  Formulation}
\label{sec: 3_Preliminaries}

\begin{figure*}[t]
\centering
\includegraphics[width=1\textwidth]{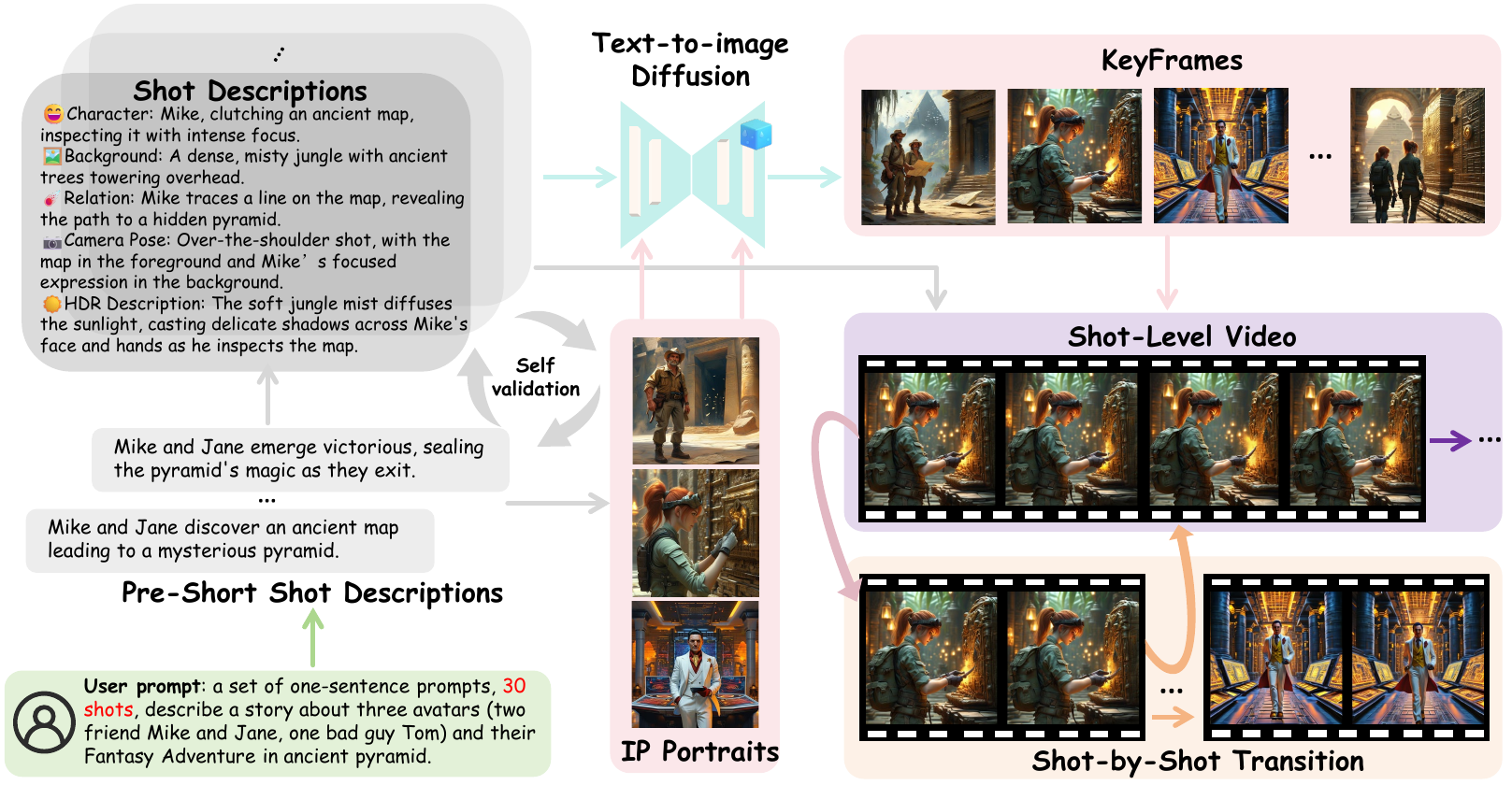}
\caption{\textbf{The FlowChart of \textit{VideoGen-of-Thought}}. \textbf{Left:} Shot descriptions are generated based on user prompts, describing various attributes such as character details, background, relations, and camera pose. Pre-shot descriptions provide a broader context for the upcoming scenes. \textbf{Middle Top:} Keyframes are generated using a text-to-image diffusion model conditioned with identity-preserving (IP) embeddings, which ensures consistent representation of characters throughout the shots. IP portraits help maintain visual identity consistency. \textbf{Right:} The shot-level video clips are generated from keyframes, followed by shot-by-shot transition inference to ensure temporal consistency across different shots. This collaborative framework ultimately produces a cohesive narrative-driven video. 
\vspace{-1em}
}
\label{fig:2}
\end{figure*}

\subsection{Preliminaries}
\label{sec:preliminaries}

Diffusion models~\cite{ho2020denoising,song2020ddim,song2020scorebased} are generative models trained to approximate data distributions $p(x)$ through iterative denoising of random noise $\epsilon \sim \mathcal{N}(0, \mathbf{I})$. The forward process gradually adds noise according to a variance schedule $\beta_t$:
\begin{equation}
q(x_t | x_{t-1}) = \mathcal{N}\left(x_t; \sqrt{1 - \beta_t} x_{t-1}, \beta_t \mathbf{I}\right)
\end{equation}
over $T$ timesteps, producing progressively noisier latents $\{x_t\}_{t=1}^T$. The reverse process learns a parameterized model $\mu_\theta$ to reconstruct $x_0$ through transitions:
\begin{equation}
p_\theta(x_{t-1} | x_t) = \mathcal{N}\left(x_{t-1}; \mu_\theta(x_t, t), \Sigma_\theta(x_t, t)\right)
\end{equation}
where $\mu_\theta$ and $\Sigma_\theta$ denote the predicted mean and variance. Training minimizes the noise prediction error via:
\begin{equation}
\mathcal{L}_{\text{uncond}} = \mathbb{E}_{x_0, \epsilon \sim \mathcal{N}(0, \mathbf{I}), t} \left[ \| \epsilon - \epsilon_\theta(x_t, t) \|^2 \right]
\label{eq:loss}
\end{equation}
Latent diffusion models~\cite{rombach2022stablediffusion} map this process to a compressed space using a VAE~\cite{kingma2013vae}, enabling conditional generation through cross-attention:
\begin{equation}
\text{Attention}(Q, K, V) = \text{Softmax}\left( \frac{Q K^\top}{\sqrt{d_k}} \right) V
\end{equation}
where $Q = W_Q\mathcal{E}(x_t)$, $K = W_K\tau_\theta(y)$, and $V = W_V\tau_\theta(y)$ for text embeddings $y$. Video diffusion models~\cite{he2022lvdm,blattmann2023stable} process frame sequences $\{z^f\}_{f=0}^{F-1} \in \mathbb{R}^{h \times w \times d}$ through temporal-aware denoising networks $\epsilon_\theta(z_t, t, c)$.
FIFO-Diffusion~\cite{kim2024fifo} extends this through queue-based latent processing:
\begin{equation}
Q_k = \{z^f\}_{f=f_k}^{f_k+n} \leftarrow \Phi(Q_k, \tau_k, c; \epsilon_\theta)
\end{equation}
where $\Phi$ denotes the DDIM sampler~\cite{song2020ddim}. While effective for frame continuity, this approach struggles with: (1) abrupt shot transitions that disrupt queue coherence, and (2) integration of image-based conditions. Our framework addresses these through reset mechanisms in adjacent latent-space processing.

\subsection{Problem Definition}
\label{sec:Problem_Definition}
Given a one-sentence user input $S$ specifying $N$ shots (\textit{e.g., "A story of Mary's life from birth to death"}), we aim to generate a multi-shot video $V$ with \textit{reasonable storylines} and \textit{visual consistency} \textbf{from one sentence} through \textbf{minimal manual intervention}. The core challenges are:

\begin{itemize}
\item \textbf{Reasonability}: Maintaining logical narrative flow across evolving storylines
    
\item \textbf{Consistency}: Preserving high-level identity~\cite{ye2023ip-adapter,yuan2024consisid,zhou2025storydiffusion} while allowing trait variations (\textit{e.g., aging expressions, contextual relationships}) across shots
    
\item \textbf{Multi-Shot Generation}: Producing minute-level videos with diverse yet interconnected shots
\end{itemize}

\textit{VideoGen-of-Thought (VGoT)} addresses these through script preparation $\{p_i\}^{N-1}_{i=0}$, identity-preserved keyframes $\{I_i\}^{N-1}_{i=0}$, and cross-shot video latents $\{z^f\}^{F-1}_{f_i=0}$. Our framework overcomes limitations in existing multi-shot methods through narrative-visual coherency mechanisms.

%% file: sec/4_Methods.tex
\section{Method: VideoGen-of-Thought}
\label{sec:method}
In this section, we introduce a structured, step-by-step framework for generating multi-shot videos with \textit{reasonable storylines} and \textit{visual consistency} \textbf{from one sentence} with \textbf{minimal manual intervention}, addressing the core challenges of narrative fragmentation, visual inconsistency, and transition artifacts through four distinct yet collaborative modules (Fig~\ref{fig:2}):


\subsection{Dynamic Storyline Modeling with Self-Vaildation}
We formulate narrative generation as constrained multi-shot decomposition with auto-regressive validation. Given user prompt $S$ and shot count $N$, our system first generates a story draft $S' = \{s_i\}_{i=1}^N$ of short shot descriptions, then produces structured scripts through:
\begin{equation}
\mathcal{P} = \{p_i\}_{i=1}^N = \bigcup_{i=1}^N \mathcal{M}_{\text{LLM}}(s_i|\mathcal{C}_{\text{film}}, \{p_j\}_{j=1}^{i-1})
\label{eq:4}
\end{equation}
where $\mathcal{C}_{\text{film}} (p_{\text{cha}}, p_b, p_r, p_{\text{cam}}, p_h)$ encodes five cinematic dimensions: $p_{\text{cha}}$ governs character appearance evolution and role relationships, $p_b$ ensures background consistency across scene transitions, $p_r$ maintains interaction patterns and event causality, $p_{\text{cam}}$ specifies shot composition through camera movements, and $p_h$ regulates HDR lighting continuity. The self-validation mechanism employs two criteria:
\begin{equation}
\mathcal{V}(p_i) = \mathbb{I}[C(p_i,p_{i-1}) > \tau_c] \cdot \mathbb{I}[K(p_i,\mathcal{C}_{\text{film}}) > \tau_k]
\end{equation}
where $C: \mathcal{P} \times \mathcal{P} \rightarrow [0,1]$ measures narrative coherence through pretrained textual feature extractor $E_{\text{GLM}}$~\cite{glm2024chatglm} to compute semantic similarity between consecutive shots, and $K: \mathcal{P} \times \mathcal{C} \rightarrow \{0,1\}$ verifies constraint completeness via rule-based checks against $\mathcal{C}_{\text{film}}$. Thresholds $\tau_c=0.85$ and $\tau_k=1$ ensure strict adherence to cinematic principles.

\begin{algorithm}[H]
\caption{Self-Validated Script Generation}
\label{alg:1}
\begin{algorithmic}[1]
\Require User prompt $S$, shot count $N$, constraints $\mathcal{C}_{\text{film}}$
\State Initialize $\mathcal{P} \gets \emptyset$, $p_{\text{prev}} \gets \emptyset$
\State Generate draft $S' \gets \{s_1,...,s_N\} = \mathcal{M}_{\text{LLM}}(S, N)$
\For{$i \gets 1$ to $N$}
    \Repeat
        \State $p_i' \gets \mathcal{M}_{\text{LLM}}(s_i, \mathcal{C}_{\text{film}}, p_{\text{prev}})$
        \State $C_{i} \leftarrow \mathbb{I}[C(p_i,p_{i-1}) > \tau_c]$
        \State $K_{i} \leftarrow \mathbb{I}[K(p_i,\mathcal{C}_{\text{film}}) > \tau_k]$
        \State $\mathcal{V}(p_i) \leftarrow C_{i} \cdot K_{i}$
    \Until{$\mathcal{V}(p_i') = 1$}
    \State $\mathcal{P} \gets \mathcal{P} \cup p_i'$, $p_{\text{prev}} \gets p_i'$
\EndFor
\end{algorithmic}
\end{algorithm}

Our dynamic storyline modeling transforms user prompts into shot sequences through a two-stage process with self-validation mechanisms that enforce narrative coherence by rejecting candidates violating cinematic principles (Algorithm~\ref{alg:1}).


\subsection{Identity-Aware Cross-Shot Propagation}
We resolve visual inconsistency through cross-shot propagation machanism, which maintains critical attributes (\textit{e.g., hairstyle, facial structure}) while permitting narrative-driven variations (\textit{e.g., expression, aging}). Using scripts $\mathcal{P}$, we generate keyframes with consistent visual identities through a two-stage process:
\begin{equation}
\mathbf{I} = \{\mathbf{I}_i\}_{i=1}^N = \mathcal{F}(\mathcal{P}, \Psi)
\end{equation}
where $\mathcal{F}$ represents our identity-preserving generation pipeline and $\Psi$ denotes the parameters of the character schema. For each shot script $p_i \in \mathcal{P}$, we extract:
\begin{equation}
e^T_i = E_{\text{GLM}}(p_i) \in \mathbb{R}^d, \quad \mathcal{C}_{\text{char}} = \mathcal{M}_{\text{LLM}}(\mathcal{P}) = \{c_j\}_{j=1}^M
\end{equation}
where $E_{\text{GLM}}$ is the text encoder and $\mathcal{C}_{\text{char}}$ contains $M$ identity descriptors (e.g., \textit{Young Mary, Elderly Mary} when describing given scripts \textit{Mary's life} ). Identity-Preserving Portrait (IPP) tokens are synthesized through:
\begin{equation}
\text{IPP}_j = \mathcal{M}_I(c_j) \in \mathbb{R}^{H \times W \times 3}, \quad e^I_j = E_{\text{CLIP}}(\text{IPP}_j) \in \mathbb{R}^{d_v}
\end{equation}
where $\mathcal{M}_I$ is a pre-trained text-to-image model and $E_{\text{CLIP}}$ denotes CLIP~\cite{radford2021clip}'s vision encoder. We inject identity features into diffusion via cross-attention:
\begin{align}
Q &= W_Q z_t \in \mathbb{R}^{n \times d_k} \\
K &= \lambda[W_K e^T_i; W'_K e^I_j] \in \mathbb{R}^{2n \times d_k} \\
V &= \lambda[W_V e^T_i; W'_V e^I_j] \in \mathbb{R}^{2n \times d_v} \\
\text{Attn}(Q,K,V) &= \text{softmax}\left(\frac{QK^\top}{\sqrt{d_k}}\right)V
\end{align}
where $z_t \in \mathbb{R}^{n \times d}$ is the latent in step $t$, $\lambda$ balances the influence of text / identity, and $[;]$ denotes concatenation. Keyframe generation integrates both modalities:
\begin{equation}
I_i = \mathcal{D}\left(z_T \big| e^T_i, e^I_{j(i)}\right), \quad j(i) = \mathcal{M}_{\text{LLM}}(p_i, \mathcal{C}_{\text{char}})
\end{equation}

By anchoring IPP tokens in narrative-derived character descriptors $\mathcal{C}_{\text{char}}$ and integrating them via attention mechanisms, we achieve robust identity fidelity across shots.

\begin{figure*}[t]
\centering
\includegraphics[width=1.0\linewidth]{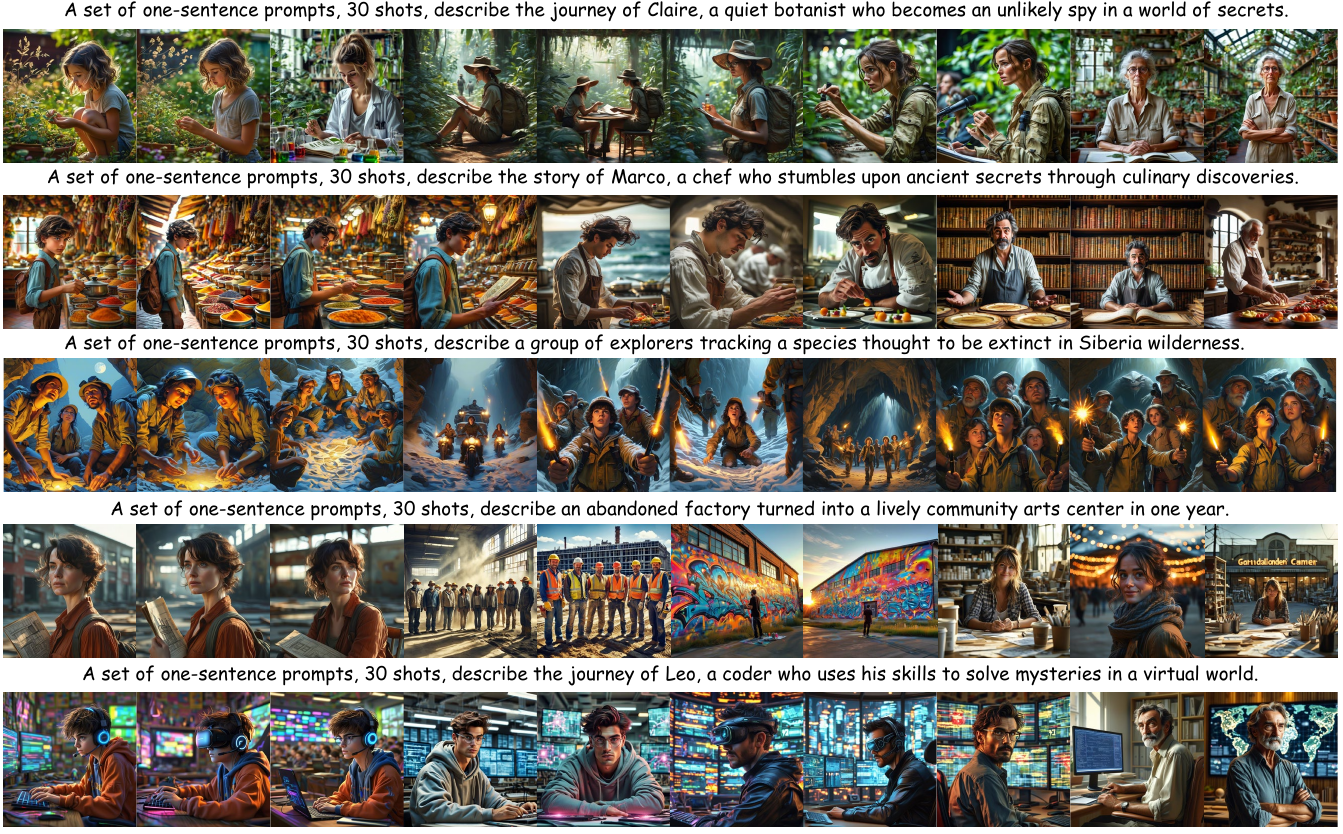}
\caption{\textbf{Visual showcases of \textit{VGoT} generated multi-shot videos.} }
\vspace{-1em}
\label{fig:4}
\end{figure*}


\subsection{Adjacent Latent Transition Mechanisms}
We address transition artifacts through latent-space noise management across shot boundaries. Given keyframes $\{I_i\}_{i=1}^N$ and script embeddings $\{e^T_i\}_{i=1}^N$, we generate shot-wise latents:
\begin{equation}
Z_i = \mathcal{M}_V(e^T_i, e^I_i, \epsilon_i) \in \mathbb{R}^{f \times c \times h \times w}
\end{equation}
where $e^T_i = E_{\text{GLM}}(s_i)$ uses simplified shot description $s_i$ rather than detailed script $p_i$, $e^I_i$ is the keyframe embedding, and $\epsilon_i \sim \mathcal{N}(0,\mathbf{I})$ is the initial noise. Inspired by FIFO~\cite{kim2024fifo}, we implement boundary-aware noise reset for cross-shot transitions:
\begin{align}
\epsilon_{\text{boundary}} &\sim \mathcal{N}(0, \beta_i\mathbf{I}), \quad \beta_i = \gamma \cdot (1 - \frac{i}{N}) \\
Z_{\text{final}} &= \mathcal{R}(Z_1, Z_2, \ldots, Z_N)
\end{align}
where $\beta_i$ controls noise magnitude at shot boundaries with scaling factor $\gamma$, and $\mathcal{R}$ denotes our reset function:
\begin{equation}
\mathcal{R}(Z_1, \ldots, Z_N) = \left[ Z_1^{1:f}, Z_2^{1:f}, \ldots, Z_N^{1:f} \right]
\end{equation}
For each transition between shots $i$ and $i+1$, we reset the diffusion process with:
\begin{equation}
z_{i+1}^0 = \epsilon_{\text{boundary}} + \alpha \cdot z_i^f
\end{equation}
where $\alpha \in [0,1]$ controls the temporal continuity and $z_i^f$ is the final latent frame of the shot $i$. The complete video generation becomes:
\begin{equation}
V = \mathcal{D}\left( \bigcup_{i=1}^N Z_i \right) \in \mathbb{R}^{T \times 3 \times H \times W}
\end{equation}

\begin{figure*}[t]
\centering
\includegraphics[width=1.0\linewidth]{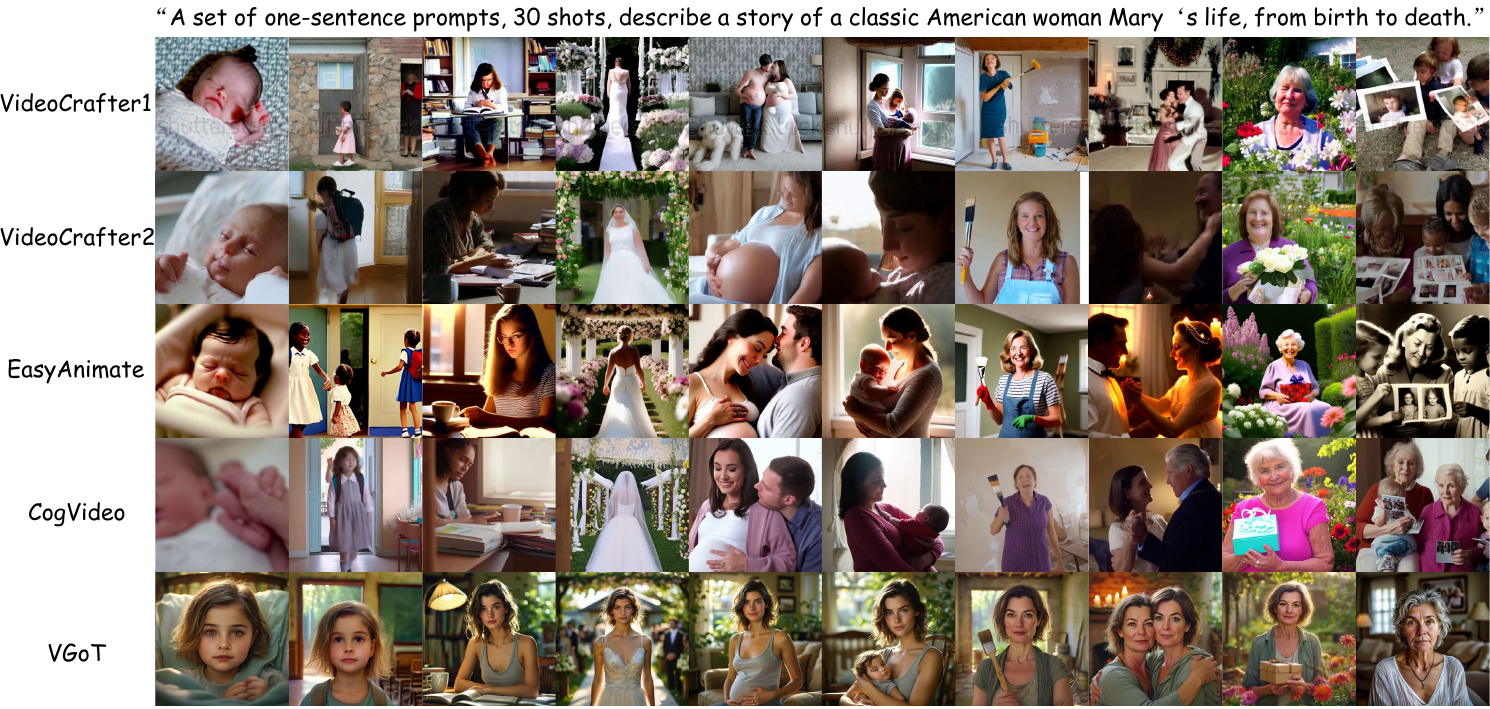}
\caption{\textbf{Visual comparison of \textit{VGoT} and baselines}}
\label{fig:3}
\end{figure*}

This mechanism decodes $\{Z_i\}_{i=1}^N$ into a coherent video $V \in \mathbb{R}^{T \times 3 \times H \times W}$ through boundary reset operations, preserving both narrative flow and visual continuity across $T = N \times f$ frames without requiring additional training.

\begin{table}[t]
  \small
  \caption{\textbf{Quantitative comparison with state-of-the-art T2V baselines.} We compare average CLIP scores, and the average FC and SC scores within and across shots between VGoT and baseline models. We use $\bold{bold}$ to highlight the highest and $\underline{underline}$ for the second high.}
  \centering
  \resizebox{1.0\textwidth}{!}{%
    \begin{tabular}{@{}lccccccccc@{}}
      \toprule
      Model & CLIP $\uparrow$ & WS-FC $\uparrow$ & CS-FC $\uparrow$ & WS-SC $\uparrow$ & CS-SC $\uparrow$ \\
      \midrule
      EasyAnimate~\cite{xu2024easyanimate} & 0.2402 & 0.4705 & 0.0268 & 0.7969 & 0.2037\\
      CogVideo~\cite{yang2024cogvideox} & 0.2477 & \underline{0.6099} & 0.0222 & 0.7424 & \underline{0.2069}\\
      VideoCrafter1~\cite{chen2023videocrafter1} & 0.2478 & 0.3706 & 0.0350 & 0.7623 & 0.1867\\
      VideoCrafter2~\cite{chen2024videocrafter2} & \underline{0.2529} & 0.5569 & \underline{0.0686} & \underline{0.7981} & 0.1798\\

      VGoT & \textbf{0.2557} & \textbf{0.8138} & \textbf{0.2688} & \textbf{0.9717} & \textbf{0.4276} \\ 
      \bottomrule
    \end{tabular}%
  }
  \raggedright
  \label{tab:1}
\end{table}

\begin{table*}[t]
  \caption{\textbf{Human Evaluation.} We compare \textit{VGoT} with baseline models in terms of Within-Shot Consistency, Cross-Shot Consistency, and Visual Quality.}
  \centering
   \resizebox{1.0\textwidth}{!}{%
  \begin{tabular}{@{}lccccccccc@{}}
    \toprule
    & \multicolumn{3}{c}{Within-Shot Consistency} & \multicolumn{3}{c}{Cross-Shot Consistency} & \multicolumn{3}{c}{Visual Quality} \\
    \cmidrule(lr){2-4} \cmidrule(lr){5-7} \cmidrule(lr){8-10}
    & Bad $\downarrow$ & Normal $\sim$ & Good $\uparrow$ & Bad $\downarrow$ & Normal $\sim$ & Good $\uparrow$ & Bad $\downarrow$ & Normal $\sim$ & Good $\uparrow$ \\
    \midrule
    EasyAnimate~\cite{xu2024easyanimate} & 0.3333 & 0.3232 & 0.3434 & 0.3535 & 0.3535 & 0.3131 & 0.4646 & 0.2727 & 0.2828 \\
    
    CogVideo~\cite{yang2024cogvideox} & 0.1341 & 0.4146 & 0.4512 & 0.2927 & 0.5976 & 0.2317 & 0.1463 & 0.4512 & 0.5244 \\    
    VideoCrafter1~\cite{chen2023videocrafter1} & 0.5446 & 0.2574 & 0.1980 & 0.6436 & 0.1881 & 0.1683 & 0.6535 & 0.1782 & 0.1683 \\
    VideoCrafter2~\cite{chen2024videocrafter2} & 0.1262 & 0.4854 & 0.3883 & 0.3495 & 0.3786 & 0.2718 & 0.1748 & 0.4951 & 0.3981 \\

    
    VGoT & 0.0889 & 0.2556 & 0.6556 & 0.0889 & 0.2444 & 0.6667 & 0.0889 & 0.2111 & 0.7000 \\
    \bottomrule
  \end{tabular}
    }
  \raggedright
  \label{tab:2}
\end{table*}

%% file: sec/5_experiments.tex
\begin{table*}[t]
  \small
  \caption{\textbf{Ablation Studies.} We evaluate the impact of removing key modules from our proposed framework. Metrics include CLIP Score, PSNR, IS, FC score, and SC score}
  \centering
\resizebox{\textwidth}{!}{%
  \begin{tabular}{@{}lccccccc@{}}
    \toprule
    & CLIP average $\uparrow$ & PSNR $\uparrow$ & IS $\uparrow$ & WS-FC $\uparrow$ & CS-FC $\uparrow$ & WS-SC $\uparrow$ & CS-SC $\uparrow$\\
    \midrule
    w/o DSM  w/o IPP & 0.1146 & 24.3265 & 7.4624 & 0.7364 & 0.1129 & 0.9406 & 0.3650 \\
    w DSM  w/o IPP & 0.1146 & 24.3265 & 7.5783 & 0.7305 & 0.1174 & 0.9471 & 0.3663 \\
    w/o DSM  w IPP & \textbf{0.1223} & 23.9228 & 7.4521 & \textbf{0.8745} & \textbf{0.3291} & 0.9486 & \textbf{0.4186} \\
    Full Method & 0.1111 & \textbf{25.7857} & \textbf{7.5194} & 0.8303 & 0.2738 & \textbf{0.9487} & 0.3859 \\
    \bottomrule
  \end{tabular}
    }
  \raggedright
    \textbf{*}{FC is denoted as Face Consistency, and SC is denoted as Style Consistency}
  \label{tab:ablation}
\end{table*}

\vspace{-1em}
\section{Experiments}


\input{sec/5-1_experiment_setting}

\input{sec/5-2_main_experiments}
\input{sec/5-3_human_evaluation}

\input{sec/5-4_ablation}

%% file: sec/5-1_experiment_setting.tex
\subsection{Experiment Settings}
\label{subsec:experiment_settings}

Current video datasets lack sufficient multi-shot narratives with consistent characters across scenes. We therefore constructed a benchmark dataset using \textit{VGoT} to create ten 30-shot stories (300 shots in total) for evaluation. Each story originates from a single user input $S$, which generates shot outlines $S'$ and detailed scripts $\mathcal{P}$ across 30 shots and five domains defined in Eq.~\ref{eq:4}. For quantitative assessment, we employ four key metrics: Within-Shot Face Consistency ($\Omega_{\text{WS-FC}}$), Cross-Shot Face Consistency ($\Omega_{\text{CS-FC}}$), Within-Shot Style Consistency ($\Omega_{\text{WS-SC}}$), and Cross-Shot Style Consistency ($\Omega_{\text{CS-SC}}$) defined in Appendix~\ref{sec:four_metrics}. We additionally report CLIP score~\cite{radford2021clip}, PSNR~\cite{fardo2016psnr}, and IS~\cite{barratt2018is_score}. Our implementation uses multiple GPT-4o~\cite{achiam2023gpt} for scripting, Kolor~\cite{kolors} as base model for keyframes, and DynamiCrafter~\cite{xing2025dynamicrafter} for video generation, compared against EasyAnimate~\cite{xu2024easyanimate}, CogVideo~\cite{hong2022cogvideo}, and VideoCrafter~\cite{chen2023videocrafter1, chen2024videocrafter2} on NVIDIA H100 GPUs.

%% file: sec/5-2_main_experiments.tex
\subsection{Comparison Evaluation}
\label{sec:experiments}
Our evaluation compares \textit{VGoT} against state-of-the-art text-to-video models using narrative scenarios from Sec~\ref{subsec:experiment_settings}. Quantitative results in Table~\ref{tab:1} demonstrate \textit{VGoT}'s superiority as follows.

\textit{VGoT} achieves \textbf{0.8138} $\Omega_{\text{WS-FC}}$ and \textbf{0.9717} $\Omega_{\text{WS-SC}}$, outperforming the best baselines (VideoCrafter2's 0.5569 $\Omega_{\text{WS-FC}}$ and 0.7981 $\Omega_{\text{WS-SC}}$) by \textbf{46.1\%} and \textbf{21.7\%} respectively. For cross-shot consistency, \textit{VGoT}'s \textbf{0.2688} $\Omega_{\text{CS-FC}}$ and \textbf{0.4276} $\Omega_{\text{CS-SC}}$ surpass VideoCrafter1's second-best 0.0686 $\Omega_{\text{CS-FC}}$ and CogVideo's 0.2069 $\Omega_{\text{CS-SC}}$. While outperform text-visual alignment via best CLIP score (\textbf{0.2557}), \textit{VGoT} maintains this performance while requiring \textbf{10$\times$ less} manual input, more qualitative analysis 


%% file: sec/5-3_human_evaluation.tex



Human evaluations (Table~\ref{tab:2}) confirm these findings: VGoT receives \textbf{66.7\%} "Good" ratings for cross-shot consistency versus 27.2\% for VideoCrafter2 and 23.2\% for CogVideo. In visual quality, 70.0\% of evaluators rate VGoT's outputs as "Good" compared to 52.4\% for CogVideo. This preference is qualitatively validated in Fig~\ref{fig:3}, which demonstrates VGoT's superior maintenance of character consistency and visual coherence across extended narratives compared to baseline outputs.

%% file: sec/5-4_ablation.tex
\subsection{Ablation Studies}
\label{subsec:ablation_studies}
We analyze two core components through systematic removal: (1) Dynamic Storyline Modeling (DSM) and (2) Identity-Preserving Portraits (IPP). Using the 30-shot cycling narrative, Table~\ref{tab:ablation} reveals three critical patterns:

1. \textbf{CLIP-Logic Tradeoff}: The full model achieves lowest CLIP score (0.1111 vs 0.1223 baseline) but highest PSNR (25.79) and IS (7.52), confirming that DSM's narrative enrichment and IPP's consistency control prioritize cinematic quality over literal prompt matching.

2. \textbf{Consistency Costs}: Removing DSM boosts $\Omega_{\text{CS-FC}}$ by 20.2\% (0.3291 vs 0.2738) and SC by 8.5\% (0.4186 vs 0.3859), but as Fig~\ref{fig:5} shows, this comes at severe narrative diversity loss—identical camera angles and repetitive scenes dominate DSM-ablated outputs.

3. \textbf{Component Synergy}: IPP alone achieves 0.8745 $\Omega_{\text{WS-FC}}$ (5.3\% higher than full model), but combined DSM+IPP delivers optimal balance—8.3\% better $\Omega_{\text{CS-SC}}$ than IPP-only versions while maintaining visual quality (25.79 PSNR vs 23.92).

These results prove both components are essential: DSM enables story progression while IPP ensures continuity, together resolving the consistency-diversity paradox in multi-shot generation.

%% file: sec/6_conclusion.tex

\section{Conclusion}
\label{sec:conclusion}

We present VideoGen-of-Thought (\textit{VGoT}), a structured framework that automates multi-shot video generation with \textit{reasonable storylines} and \textit{visual consistency} \textbf{from one sentence} through \textbf{minimal manual intervention}, through three core innovations: dynamic storyline modeling, identity-aware cross-shot propagation, and adjacent latent transition mechanisms.
\textit{VGoT} achieves 20.39\% higher within-shot face consistency and 17.36\% better style consistency than previous state-of-the-art methods, with 10× fewer manual adjustments than alternatives while maintaining director-level narrative flow.
Our work redefines automated multi-shot generation, bridging raw visual synthesis with cinematic storytelling through systematic visual stories decomposition.

%% file: sec/X_suppl.tex
\clearpage
\setcounter{page}{1}

\section{Limitations, Licenses, and Future Work}
\label{sec:limits_licenses_future}

\paragraph{Limitations.}
VGoT relies on pretrained components without additional finetuning, which bounds performance by the base models' capabilities. In particular, DynamiCrafter~\cite{xing2025dynamicrafter} may limit motion diversity under highly complex camera trajectories, reduce temporal stability in out-of-distribution scenes with rapid appearance changes, and constrain very long-range dependencies beyond adjacent-shot transitions. These constraints are characteristic of training-free pipelines and motivate future model-level improvements.

\paragraph{Licenses Declaration.}
We acknowledge and comply with licenses of third-party assets used in our pipeline. DynamiCrafter is distributed under the Apache License 2.0. Kolor is distributed under the Apache-2.0 license. We use these tools within their permitted scopes and cite their sources. Other services used for scripting (e.g., commercial LLM APIs) are accessed under their respective terms of service. Our released code and evaluation scripts will clearly indicate all external dependencies and their licenses.

\paragraph{Future Work.}
We plan to: (1) integrate stronger video backbones and optional finetuning to enhance motion diversity and long-range temporal reasoning;
(2) extend identity handling to multi-subject IPP with fine-grained attribute disentanglement; 
(3) broaden cultural and linguistic coverage in prompts and benchmarks; 
(4) include optional professional and sturctured movie screenplay writing in the script generation process.

\section{Detailed Example of Dynamic Storylines}
\label{sec:dynamic_storyline}

\textit{Dynamic Storylines Modeling} plays a fundamental role in converting high-level user input into a detailed series of prompts for each shot within the multi-shot video generation process. The specific process is to convert a single sentence user input $S$, into a more detailed and structured description $S'$, which is then decomposed into a set of prompts \( P = \{p_1, p_2, \dots, p_N\} \), corresponding to each of the \( N \) shots required for the complete video. This process uniformly adopts a large language model (LLM) and uses prompt engineering to ensure the reasonableness of the generated video. This process enables the generation of a logical, stepwise narrative structure that serves as the backbone for subsequent keyframe and shot-level generation.

For example, consider the user input: \textit{e.g., a set of one-sentence prompts, 30 shots, describe a story about three avatars (two friend Mike and Jane, one bad guy Tom) and their fantasy adventure in an ancient pyramid.}. The LLM would first transform this input into a detailed version $S'$ consisting of 30 one-sentence scripts like $s_1$: \textit{``Mike’s Discovery: Mike examines an ancient map with intense focus, revealing the path to the pyramid.''} Subsequently, each $s_{i}$ in $S'$ is decomposed into prompts $P$, such as $p_1$: \textit{\textbf{Character}: Mike, holding an ancient map with Jane by his side. \textbf{Background}: A dense jungle filled with mist and towering trees. \textbf{Relation}: Mike studies the map closely, pointing to a pyramid. \textbf{Camera Pose}: Medium shot focusing on Mike and Jane. \textbf{HDR Description}: Soft light filters through the trees, creating dynamic shadows on the map and characters."}. This structured approach ensures narrative coherence across multiple shots, laying the foundation for the generation process.

Our self-validation mechanism ensures cinematic rigor through iterative refinement. Consider a draft shot description: \textit{2. Inside the helicopter, a diverse team of scientists, their faces filled with anticipation and anxiety, pore over maps and equipment}

\textbf{Initial Generation Attempt:}
\begin{equation*}
p_2^{(1)} = \begin{cases} 
p_{\text{cha}} & \text{Team of scientists with mixed ages} \\
p_b & \text{Helicopter interior} \\
p_r & \text{Studying maps} \\
p_{\text{cam}} & \text{Close-up} \\
p_h & \text{(Missing HDR specification)}
\end{cases}
\end{equation*}

Validation fails with $K(p_2^{(1)}, \mathcal{C}_{\text{film}}) = 0$ due to incomplete HDR description. The LLM regenerates with lighting constraints:

\textbf{Validated Output:}
\begin{equation*}
p_2^{*} = \begin{cases} 
p_{\text{cha}} & \text{Diverse team in expedition gear, anxious expressions} \\
p_b & \text{Helicopter over Amazon rainforest} \\
p_r & \text{Team engrossed in equipment} \\
p_{\text{cam}} & \text{Wide shot showing interior/exterior} \\
p_h & \text{Sunlight streaming through windows, warm glow}
\end{cases}
\end{equation*}

This satisfies both criteria: $C(p_2^{*},p_1) = 0.92 > \tau_c$, $K(p_2^{*},\mathcal{C}_{\text{film}}) = 1$, ensuring director-level coherence from user input under minimal manual intervention.

\section{Multi-Shot Evaluation Protocol}
\label{sec:four_metrics}
Existing evaluation frameworks predominantly focus on single-shot quality, leaving multi-shot assessment underspecified. We introduce a hierarchical protocol that quantifies both intra-shot and inter-shot consistency:
\begin{equation}
\Omega_{\text{multi-shot}} = \{\Omega_{\text{WS-FC}}, \Omega_{\text{CS-FC}}, \Omega_{\text{WS-SC}}, \Omega_{\text{CS-SC}}\}
\end{equation}

\noindent\textbf{Within-Shot Face Consistency (WS-FC)} measures identity preservation within temporal sequences:
\begin{equation}
\Omega_{\text{WS-FC}}(V_i) = \frac{1}{f-1} \sum_{j=1}^{f-1} \cos\langle F_j^i, F_{j+1}^i \rangle
\end{equation}
where $F_j^i$ represents facial features from frame $j$ of shot $i$ extracted using InsightFace~\cite{ren2023pbidr, guo2021sample, gecer2021ostec, an_2022_pfc_cvpr, an_2021_pfc_iccvw, deng2020subcenter, Deng2020CVPR, guo2018stacked, deng2018menpo, deng2018arcface}.

\noindent\textbf{Cross-Shot Face Consistency (CS-FC)} evaluates identity preservation across shots:
\begin{equation}
\Omega_{\text{CS-FC}}(V) = \frac{1}{N-1} \sum_{i=1}^{N-1} \frac{1}{n} \sum_{j=1}^{n} \cos\langle F_{f-j+1}^i, F_{j}^{i+1} \rangle
\end{equation}
where $n=8$ in our implementation, comparing the last $n$ frames of shot $i$ with the first $n$ frames of shot $i+1$.


\begin{figure}[t]
\centering
\includegraphics[width=1.0\linewidth]{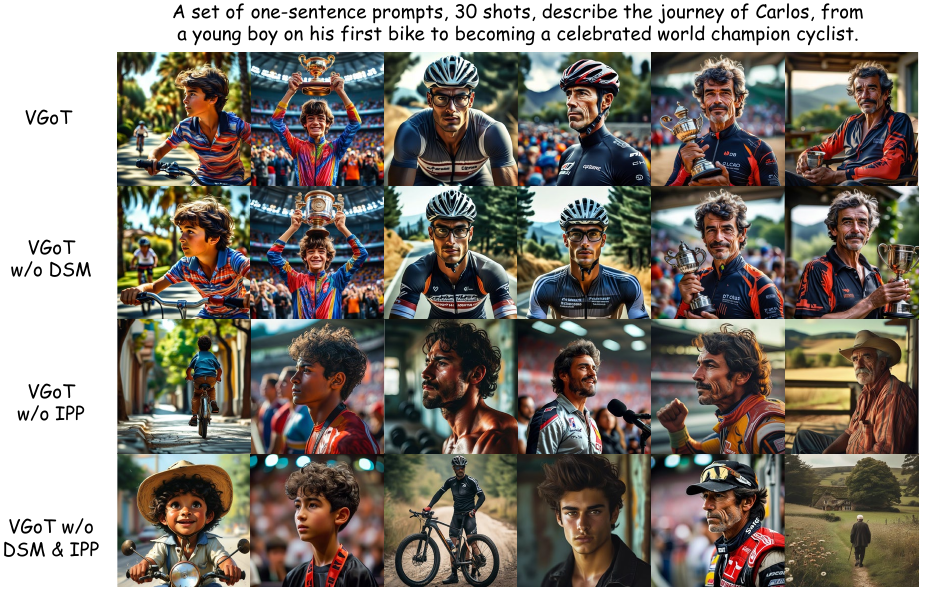}
\caption{Visual Demonstration of the ablation studies of \textbf{VGoT}}
\label{fig:5}
\end{figure}

\noindent\textbf{Within-Shot Style Consistency (WS-SC)} quantifies stylistic coherence through VGG-19~\cite{simonyan2014deep} features:
\begin{equation}
\Omega_{\text{WS-SC}}(V_i) = \frac{1}{f-1} \sum_{j=1}^{f-1} \cos\langle S_j^i, S_{j+1}^i \rangle
\end{equation}
where $S_j^i$ represents flattened VGG-19~\cite{simonyan2014deep} features from frame $j$ of shot $i$.

\noindent\textbf{Cross-Shot Style Consistency (CS-SC)} assesses style preservation between adjacent shots:
\begin{equation}
\Omega_{\text{CS-SC}}(V) = \frac{1}{N-1} \sum_{i=1}^{N-1} \frac{1}{n} \sum_{j=1}^{n} \cos\langle S_{f-j+1}^i, S_{j}^{i+1} \rangle
\end{equation}
These metrics establish a comprehensive framework capturing both local and global consistency aspects essential for multi-shot video assessment, addressing limitations in current evaluation approaches.

\begin{figure*}[!t]
\centering
\includegraphics[width=0.9\linewidth]{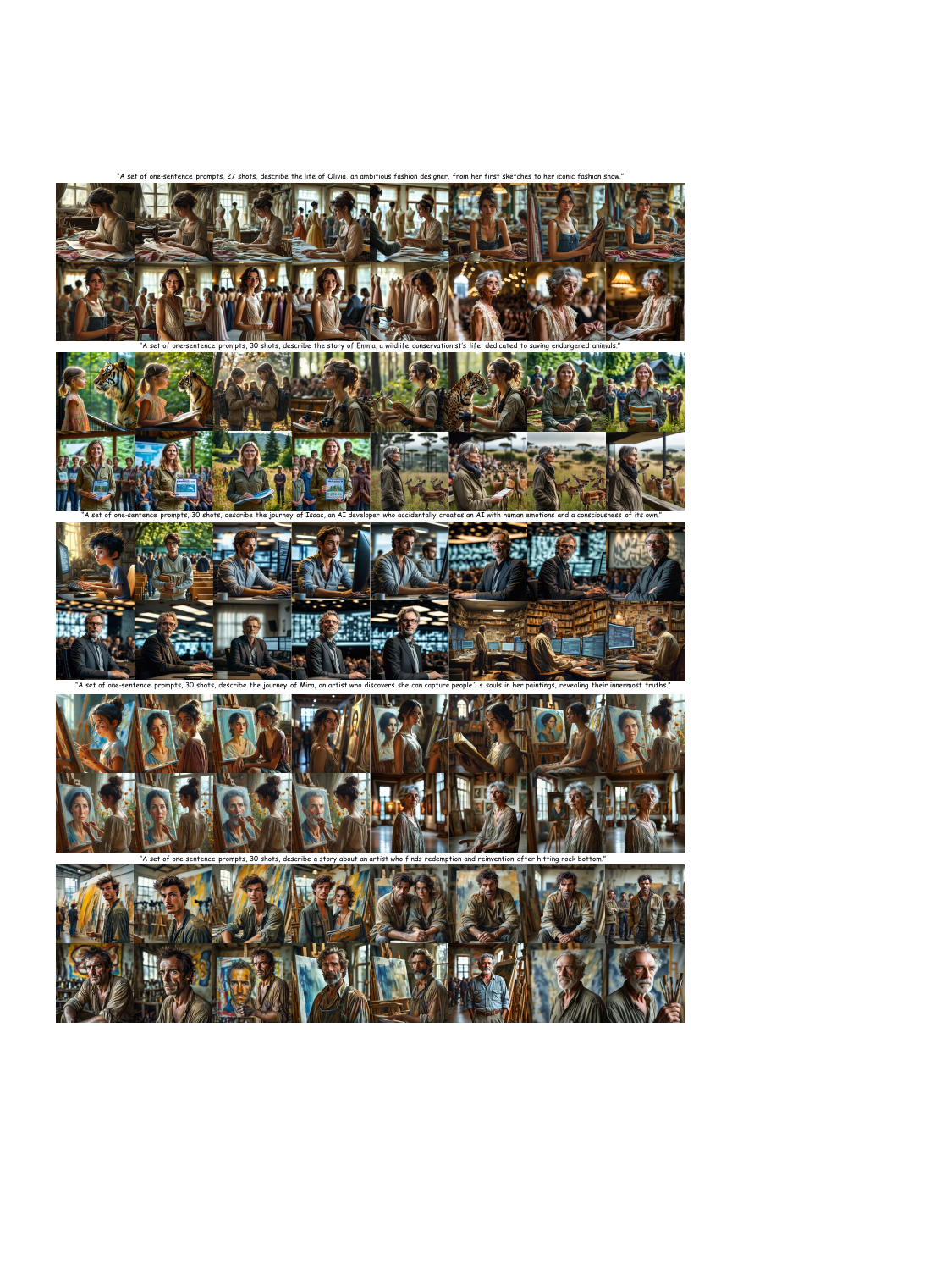}
\caption{\textit{VGoT} Visual complement of the multi-camera video generated.}
\label{VGoT:results}
\end{figure*}

\begin{figure*}[!t]
\centering
\includegraphics[width=0.9\linewidth]{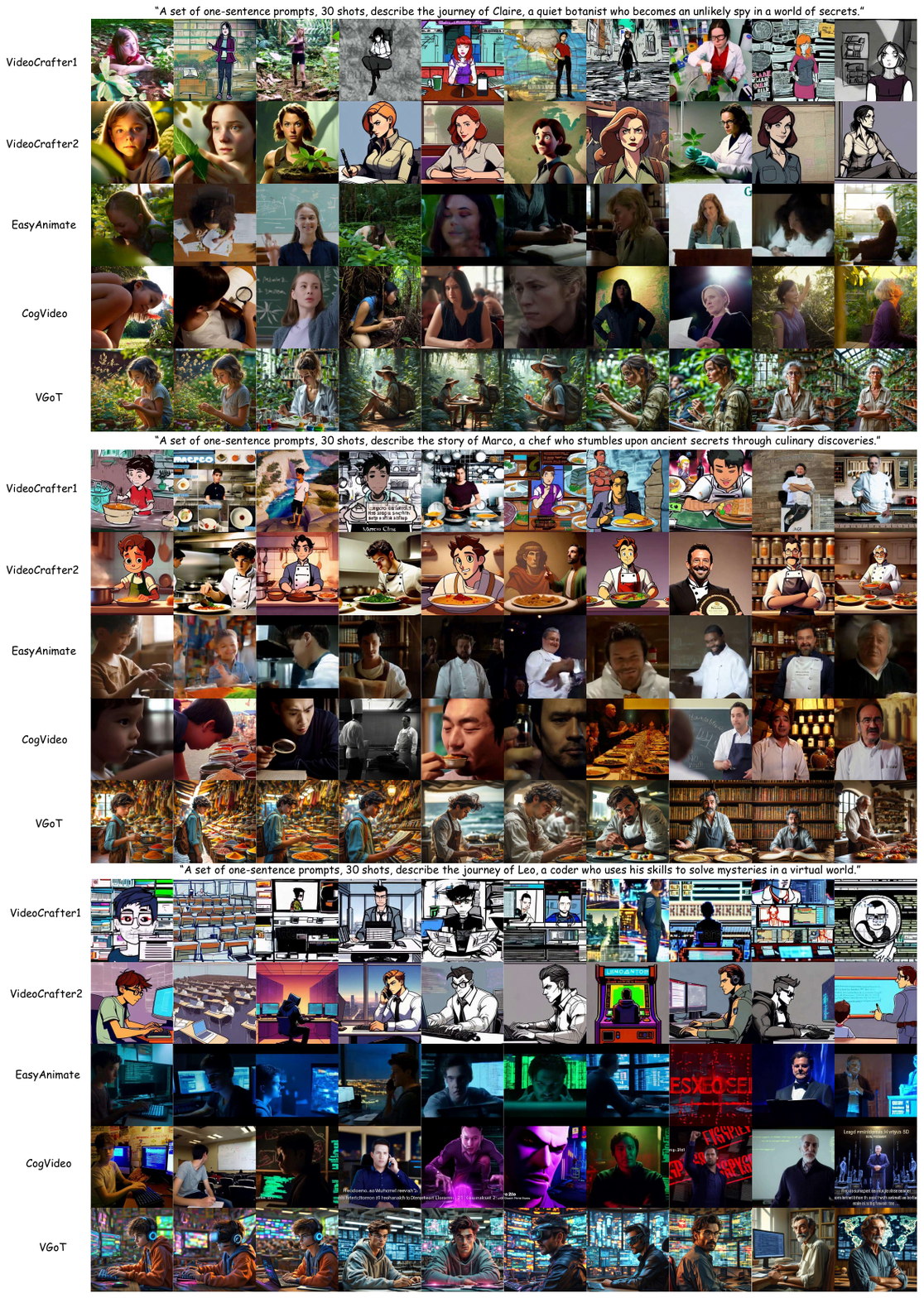}
\caption{Visual comparison of \textit{VGoT} with baselines Supplement.}
\label{VGoT_vs}
\end{figure*}

\begin{figure*}[!t]
\centering
\includegraphics[width=1.0\linewidth]{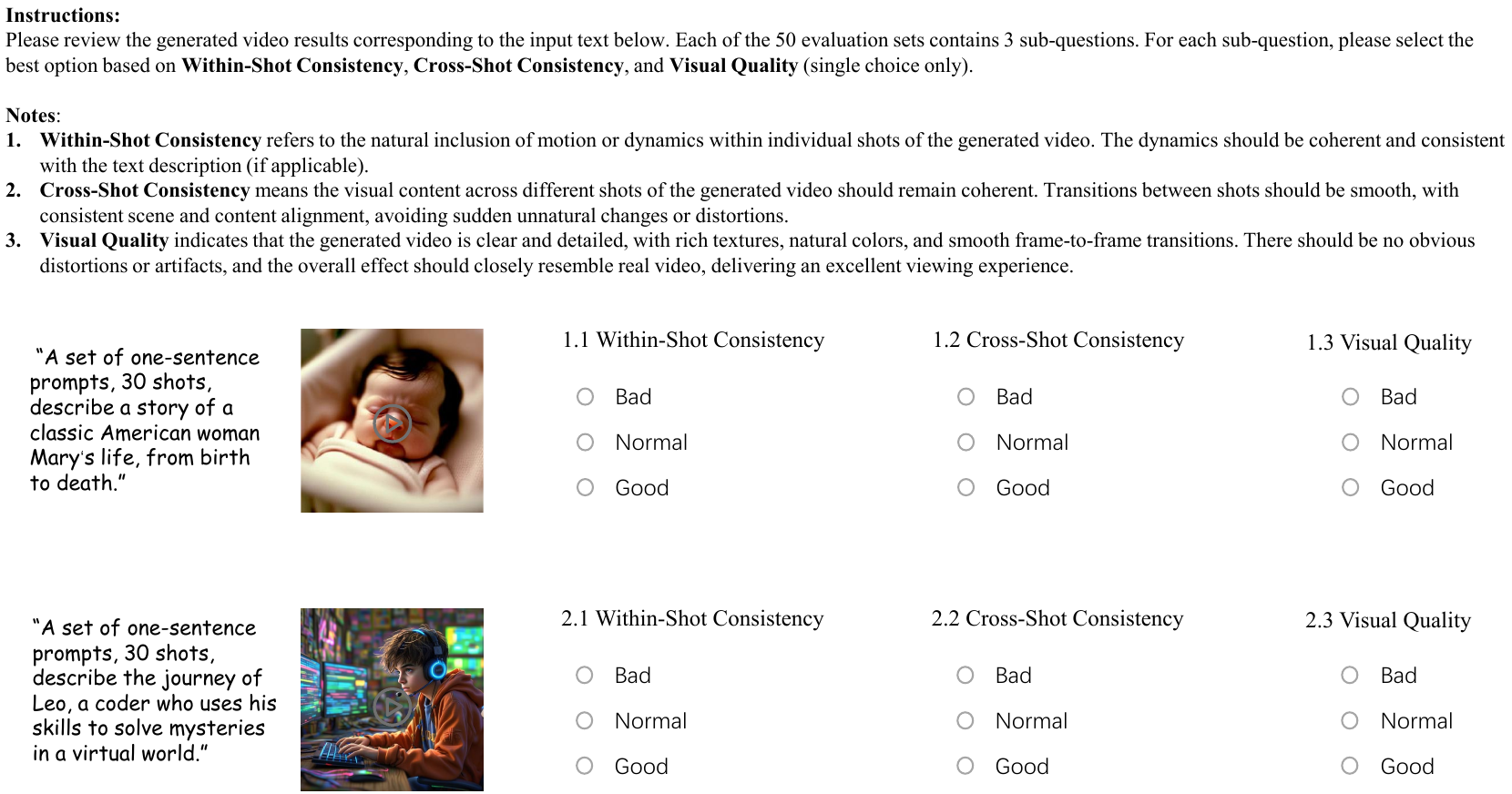}
\caption{Designed user study interface. Each participant was required to rate 50 videos by answering three sub-questions for each video. Due to page limitations, only two videos are shown here.}
\label{fig:user_study}
\end{figure*}

\section{Additional Results}
\label{sec:additional}

We validate \textit{VideoGen-of-Thought (VGoT)} through four narrative archetypes. \textbf{Type 1: Longitudinal Character Development} demonstrates decade-spanning consistency using prompts like \textit{``30-shot story of Marco discovering ancient secrets through culinary journeys"}, where our framework maintains consistency across aging sequences. \textbf{Type 2: Multi-Actor Scenes} handles complex group dynamics in scenarios like \textit{``Abandoned factory transformed into community art center"}, preserving relationship continuity between complex stories and multiple characters across shots through identity-aware propagation.
\textbf{Type 3: Non-Human Narratives} extends identity preservation to fantastical subjects, as shown in \textit{``Mycelium networks and mechanical bees restoring ecosystems"}, we explore the creatity of \textit{VGoT} in marvelous entities. We also evaluate the capability of \textit{VGoT} to create diverse stories with the same input: \textit{``an immigrant's story of moving to a new country, struggling, and eventually finding success as an entrepreneur."} in \textbf{Type 4} showcases.

Moreover, we also provide additional comparison results to illustrate \textit{VGoT}'s advantages over existing state-of-the-art methods. These comparisons include visual comparisons with four baselines: \textit{EasyAnimate}, \textit{CogVideo}, \textit{VideoCrafter1}, and \textit{VideoCrafter2}. Each comparative example is analyzed in terms of visual consistency, narrative coherence, and overall quality. As shown in Figure~\ref{VGoT_vs}, \textit{VGoT} consistently outperforms the baselines in terms of character continuity, background stability, and logical flow across shots. These results highlight \textit{VGoT}'s ability to maintain coherent storytelling while also achieving high-quality visuals.
We also prepared the original experiment data record for quantitative evaluation and ablation studies in our provided materials.

\section{User Study}
To evaluate the user-perceived quality of videos generated by our \textit{VGoT} framework, we conducted an extensive user study involving 10 participants. The participants were given 50 accelerated multi-shot videos, each generated either by \textit{VGoT} or one of four baseline methods. The 10 input stories, consisting of 30 shots each, were randomly assigned to ensure diverse feedback and minimize bias. Each user was presented with 10 videos from different sources and asked to evaluate them on a scale of \textit{good}, \textit{normal}, or \textit{bad}, based on three specific criteria: within-shot consistency, cross-shot consistency, and visual quality.

The results of the user study are summarized in Figure~\ref{fig:user_study}. The data indicates that users significantly preferred the videos generated by \textit{VGoT}, especially regarding cross-shot consistency. Users found that \textit{VGoT}'s videos maintained logical transitions between shots and preserved character appearances across different scenes, reflecting the robustness of our approach. Compared to the baselines, \textit{VGoT}'s results were rated highly for narrative coherence and overall quality, demonstrating the effectiveness of our collaborative multi-shot framework in meeting user preferences.

\section{Ethics Statement}
\label{app:ethics_statement}

\paragraph{Potential Harms Caused by the Research Process.}
Our study uses publicly available pretrained systems for scripting and generation (e.g., GPT-4o for script preparation, Kolor for keyframes, and DynamiCrafter for video synthesis) in accordance with their licenses and terms of service (see Section~\ref{subsec:experiment_settings}). Computation for metric evaluation and visual synthesis was performed on NVIDIA H100 GPUs. The 10-story benchmark used for evaluation is created by the authors specifically for this work; it does not contain personal data or copyrighted media beyond model outputs generated under the respective tools' usage policies. A small human evaluation with 10 participants was conducted to assess perceived quality and consistency (Figure~\ref{fig:user_study}); participants were informed of the study purpose, their privacy was protected, and compensation followed local norms. No sensitive personal information was collected, and we identified no additional risks to participants.

\paragraph{Societal Impact and Potential Harmful Consequences.}
VGoT is a training-free pipeline that automates multi-shot video generation from a single sentence. While this can benefit creative workflows and prototyping, risks remain. First, the environmental footprint of generative pipelines is non-negligible; H100-class accelerators consume substantial energy during inference and evaluation. Second, synthetic videos may be misused for disinformation or to fabricate misleading content if deployed irresponsibly. Third, bias can arise from scriptwriting and scene conventions (e.g., English-centric prompts or specific cultural settings), potentially reducing representativeness across regions or languages. Future work should prioritize energy-aware configurations, content provenance indicators, and broader cultural coverage in story prompts and evaluation scenarios.

\paragraph{Impact Mitigation Measures.}
We intend to release the VGoT code and evaluation scripts under an open-source license for academic research use, documented to clarify intended use and discourage misuse. We recommend that downstream deployments include visible AI-generation disclosures and optional watermarking, follow model and dataset licenses, and avoid use cases that could harm individuals or communities. We will maintain the released materials and welcome community feedback to improve responsible usage and coverage.

%% file: sec/checklist.tex
\section*{NeurIPS Paper Checklist}

\begin{enumerate}

\item {\bf Claims}
    \item[] Question: Do the main claims made in the abstract and introduction accurately reflect the paper's contributions and scope?
    \item[] Answer: \answerYes{}
    \item[] Justification: Our abstract and introduction state the three core contributions and reported gains, which match the methods in Section~\ref{sec:method} and our experimental evidence in Sections~\ref{subsec:experiment_settings} and \ref{sec:four_metrics}.

\item {\bf Limitations}
    \item[] Question: Does the paper discuss the limitations of the work performed by the authors?
    \item[] Answer: \answerYes{}
    \item[] Justification: Appendix~\ref{sec:limits_licenses_future} details our limitations tied to base model capacity and discusses future directions.

\item {\bf Theory assumptions and proofs}
    \item[] Question: For each theoretical result, does the paper provide the full set of assumptions and a complete (and correct) proof?
    \item[] Answer: \answerNA{}
    \item[] Justification: We present a systematic problem--solution framework for multi-shot video generation in Section~\ref{sec:method} and empirical evaluation protocol in Appendix~\ref{sec:four_metrics}; we do not include formal theorems or proofs that require explicit assumption sets.

    \item {\bf Experimental result reproducibility}
    \item[] Question: Does the paper fully disclose all the information needed to reproduce the main experimental results of the paper to the extent that it affects the main claims and/or conclusions of the paper (regardless of whether the code and data are provided or not)?
    \item[] Answer: \answerYes{}
    \item[] Justification: Core settings (10 stories × 30 shots), baselines, models, hardware, and metrics are specified in Sections~\ref{subsec:experiment_settings} and \ref{sec:four_metrics}; we also state intent to release code and evaluation scripts to facilitate exact reproduction (Appendix~\ref{app:ethics_statement}).

\item {\bf Open access to data and code}
    \item[] Question: Does the paper provide open access to the data and code, with sufficient instructions to faithfully reproduce the main experimental results, as described in supplemental material?
    \item[] Answer: \answerYes{}
    \item[] Justification: We intend to release the VGoT code and evaluation scripts for academic research with documentation and licensing guidance (Appendix~\ref{app:ethics_statement}).

\item {\bf Experimental setting/details}
    \item[] Question: Does the paper specify all the training and test details (e.g., data splits, hyperparameters, how they were chosen, type of optimizer, etc.) necessary to understand the results?
    \item[] Answer: \answerYes{}
    \item[] Justification: We specify story construction (10 stories, 30 shots each), baselines, models (GPT-4o, Kolor, DynamiCrafter), hardware (H100), and metrics; our pipeline is training-free, so training hyperparameters are not applicable (Section~\ref{subsec:experiment_settings}).

\item {\bf Experiment statistical significance}
    \item[] Question: Does the paper report error bars suitably and correctly defined or other appropriate information about the statistical significance of the experiments?
    \item[] Answer: \answerNo{}
    \item[] Justification: Despite we conduct quantitative comparison in Table~\ref{tab:1} and qualitative comparison in Figure~\ref{fig:3}, human evaluation in Table~\ref{tab:2}, and ablation study in Table~\ref{tab:ablation} and Figure~\ref{fig:5}, we don't report point estimates without error bars, confidence intervals, or statistical tests.

\item {\bf Experiments compute resources}
    \item[] Question: For each experiment, does the paper provide sufficient information on the computer resources (type of compute workers, memory, time of execution) needed to reproduce the experiments?
    \item[] Answer: \answerYes{}
    \item[] Justification: We disclose compute resources (H100 GPUs) and a training-free inference pipeline; experiments are reproducible under the specified story set and metrics, with compute bounded by inference-time usage (Section~\ref{subsec:experiment_settings}).
    
\item {\bf Code of ethics}
    \item[] Question: Does the research conducted in the paper conform, in every respect, with the NeurIPS Code of Ethics \url{https://neurips.cc/public/EthicsGuidelines}?
    \item[] Answer: \answerYes{}
    \item[] Justification: We use synthetic prompts and public pretrained models and do not process personal data; we adhere to standard research ethics and anonymization for submission.

\item {\bf Broader impacts}
    \item[] Question: Does the paper discuss both potential positive societal impacts and negative societal impacts of the work performed?
    \item[] Answer: \answerYes{}
    \item[] Justification: Our Ethics Statement (Appendix~\ref{app:ethics_statement}) discusses potential positive uses and negative societal impacts (energy footprint, misuse risks, and cultural bias) and outlines mitigation measures.
    
\item {\bf Safeguards}
    \item[] Question: Does the paper describe safeguards that have been put in place for responsible release of data or models that have a high risk for misuse (e.g., pretrained language models, image generators, or scraped datasets)?
    \item[] Answer: \answerNA{}
    \item[] Justification: We do not release high-risk models or scraped datasets; our paper does not introduce assets that require special safeguards.

\item {\bf Licenses for existing assets}
    \item[] Question: Are the creators or original owners of assets (e.g., code, data, models), used in the paper, properly credited and are the license and terms of use explicitly mentioned and properly respected?
    \item[] Answer: \answerYes{}
    \item[] Justification: Appendix~\ref{sec:limits_licenses_future} declares licenses for DynamiCrafter (Apache License 2.0) and Kolor (Apache-2.0) and notes compliance with external services' terms.

\item {\bf New assets}
    \item[] Question: Are new assets introduced in the paper well documented and is the documentation provided alongside the assets?
    \item[] Answer: \answerYes{}
    \item[] Justification: We will introduce new pipeline code and evaluation scripts for VGoT and will provide documentation and license notes upon release (See Appendix~\ref{app:ethics_statement}).

\item {\bf Crowdsourcing and research with human subjects}
    \item[] Question: For crowdsourcing experiments and research with human subjects, does the paper include the full text of instructions given to participants and screenshots, if applicable, as well as details about compensation (if any)? 
    \item[] Answer: \answerYes{}
    \item[] Justification: Our Ethics Statement summarizes participant privacy protection and compensation norms; Figure~\ref{fig:user_study} illustrates the interface. We provide the instruction in the supplemental material.

\item {\bf Institutional review board (IRB) approvals or equivalent for research with human subjects}
    \item[] Question: Does the paper describe potential risks incurred by study participants, whether such risks were disclosed to the subjects, and whether Institutional Review Board (IRB) approvals (or an equivalent approval/review based on the requirements of your country or institution) were obtained?
    \item[] Answer: \answerNA{}
    \item[] Justification: Our human-subject study involved rating multi-shot video quality and consistency, which is generally considered minimal risk. Participants were informed about the evaluation task; as detailed in Appendix~\ref{app:ethics_statement}, we protected privacy and identified no additional risks. We do not provide formal IRB approval details, which can be common for minimal-risk studies depending on institutional policies, but we followed ethical considerations regarding compensation and privacy.

\item {\bf Declaration of LLM usage}
    \item[] Question: Does the paper describe the usage of LLMs if it is an important, original, or non-standard component of the core methods in this research? Note that if the LLM is used only for writing, editing, or formatting purposes and does not impact the core methodology, scientific rigorousness, or originality of the research, declaration is not required.
    \item[] Answer: \answerYes{}
    \item[] Justification: LLM usage is a core component of our method and is described in Section~\ref{sec:method} (dynamic storyline modeling) and Section~\ref{subsec:experiment_settings}.

\end{enumerate}